\documentclass[letterpaper]{article} 
\usepackage{aaai2027}  
\usepackage[hyphens]{url}  
\usepackage{graphicx} 

\usepackage{booktabs}
\usepackage{multirow}
\usepackage{graphicx}
\usepackage{array}
\usepackage{tabularx}
\usepackage{amsmath}
\usepackage{amssymb}
\usepackage{cuted}
\usepackage{capt-of}

\usepackage{natbib}  
\usepackage{caption} 
\usepackage{algorithm}
\usepackage{algorithmic}

\usepackage{newfloat}
\usepackage{listings}
\DeclareCaptionStyle{ruled}{labelfont=normalfont,labelsep=colon,strut=off} 
\floatstyle{ruled}
\newfloat{listing}{tb}{lst}{}
\floatname{listing}{Listing}

\usepackage{booktabs}

\title{Motion Beyond Morphology: Bootstrapping Cross-Category Motion Transfer from Abstract Motion Representations}
\author {
    Zhixue Fang\textsuperscript{\rm 1,*},
    Zhimin Zhang\textsuperscript{\rm 2,\rm 1,*},
    Bi'an Du\textsuperscript{\rm 2,\rm 1},
    Zijie Meng\textsuperscript{\rm 2,\rm 1},
    Yan Zhou\textsuperscript{\rm 1,\ensuremath{\dagger}},
    Wei Hu\textsuperscript{\rm 2,\ensuremath{\dagger}},\\
    Guoxin Zhang\textsuperscript{\rm 1},
    Pengfei Wan\textsuperscript{\rm 1},
    Kun Gai\textsuperscript{\rm 1}
}
\affiliations {
    \textsuperscript{\rm 1}Kling Team, 
    \textsuperscript{\rm 2}Peking University\\
}

\begin{document}

\makeatletter

\setlength{\titlebox}{0pt}

\twocolumn[{
\@maketitle
\vspace{0.5em}
\begin{minipage}{\textwidth}
  \centering
  \includegraphics[width=\textwidth]{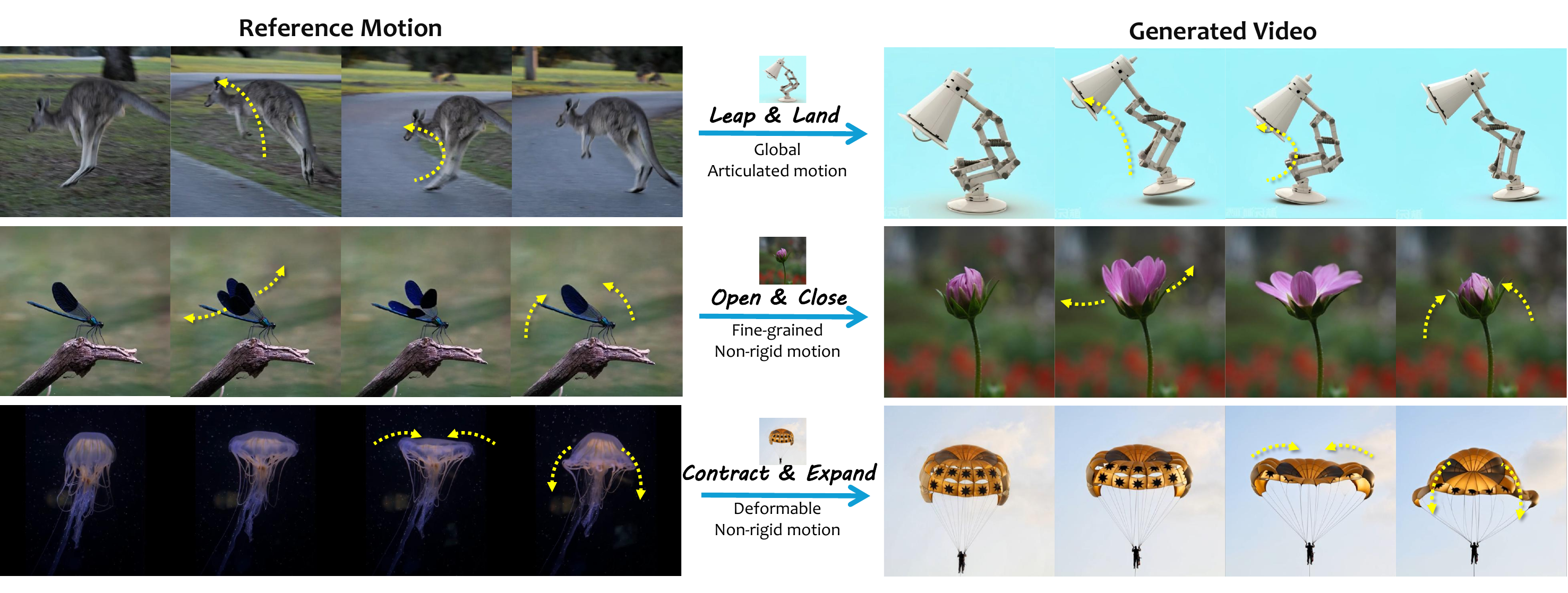}
  \captionof{figure}{
    \textbf{Motion transfer beyond morphological correspondence.}
    Our method transfers motion across substantially different subjects and structures while preserving the target appearance.
    From global articulated \emph{leap-and-land} motion
    (kangaroo $\rightarrow$ lamp), to fine-grained
    \emph{open-and-close} dynamics (dragonfly wings $\rightarrow$ flower petals), and deformable \emph{contract-and-expand} motion
    (jellyfish $\rightarrow$ parachute), the generated videos reproduce
    the underlying motion pattern despite weak or absent part-level
    correspondence. Yellow arrows visualize the transferred dynamics.
  }
  \label{fig:teaser}
\end{minipage}
}]


\begingroup
\renewcommand{\thefootnote}{\fnsymbol{footnote}}

\footnotetext[1]{\small
These authors contributed equally to this work.}

\footnotetext[2]{\small
Corresponding author.}

\endgroup

\makeatother

\begin{abstract}
Video motion transfer aims to animate a target object using dynamics from a reference video. Existing formulations largely rely on fixed structural correspondence, which becomes ill-defined when reference and target objects differ substantially in morphology, articulation, or deformation mechanisms. We introduce \textbf{Motion Beyond Morphology}, a perspective that seeks to transfer motion beyond fixed structural correspondence, by preserving dynamics that remain meaningful across different target morphologies. To realize this, we propose a two-stage framework. Stage~I learns complementary multi-granularity abstract motion views and uses them to bootstrap cross-category video pairs that preserve transferable dynamics across diverse morphologies. Stage~II internalizes this supervision into direct reference-video-conditioned generation, removing the need for explicit motion extraction at inference. We further introduce \textbf{OpenVMT-Dataset} and \textbf{OpenVMT-Bench} for training and evaluating image- and text-conditioned motion transfer across Same, Near, and Far category gaps. Extensive experiments demonstrate state-of-the-art motion fidelity and target preservation. 
Project page: {\textcolor{blue}{https://miniz233.github.io/MotionBeyondMorphology/}}
\end{abstract}

\section{Introduction}
\label{sec:introduction}

Video motion transfer aims to animate a target object according to the dynamics of a source video while preserving the appearance and scene specified by the target condition. Existing approaches are predominantly developed for source-target pairs with compatible structural templates. Human and character animation, for instance, commonly relies on pose or skeleton sequences, where source and target share comparable articulated structures and kinematic layouts~\cite{xu2023magicanimate,hu2024animateanyone}. Beyond human-centric animation, existing methods often extend to other domains by adopting representations tailored to specific structural patterns or motion regimes. While effective within their intended settings, such formulations remain difficult to generalize when source and target exhibit substantially different morphologies or motion mechanisms.

Recent studies broaden the scope of motion transfer to reduce such category-specific assumptions through implicit and explicit motion representations. Implicit features derived from intermediate activations or attention patterns encode rich spatiotemporal dynamics and offer broad category applicability~\cite{ling2024motionclone,ditflow,det}, yet may retain source-specific appearance, geometry, or background information. Explicit representations provide stronger separation from visual content, but typically capture only selected aspects of motion. Object trajectories capture object-level motion, such as global translation and displacement~\cite{wang2024motionctrl, wu2024draganything}, but provide limited information about internal articulation and deformation. Dense point tracks offer finer-grained local motion cues by tracing point-wise dynamics, yet lack explicit semantic, pose, and topological structure~\cite{geng2025motion}. Consequently, different representations offer complementary strengths, and no single representation remains equally expressive and reliable across diverse object categories and motion patterns.

These limitations motivate a broader formulation of \textbf{open-category motion transfer}, where source and target may differ substantially in semantics, morphology, and motion topology. We argue that motion transfer across open categories should not rely on a fixed notion of correspondence. Instead, the model should adaptively determine which motion factors should remain coupled between source and target and which should be disentangled according to their structural compatibility. For closely related categories, semantic correspondence, pose, and topology may provide meaningful constraints together with trajectory and temporal dynamics. As the morphological gap increases, transferable information may shift toward more abstract factors such as trajectory, rhythm, and local deformation, while source-specific semantics and structure should be progressively weakened. The relevant combination can vary across individual source--target pairs, making adaptive coupling a key requirement for general motion transfer. We refer to this perspective as \textbf{Motion Beyond Morphology}, where motion is transferred through the factors that remain meaningful under the target morphology rather than through a fixed structural template. Figure~\ref{fig:teaser} illustrates such transfers across large morphological gaps.

Learning this adaptive correspondence presents a distinct supervision challenge. Raw videos jointly encode motion, appearance, geometry, and spatial layout, while large-scale video collections rarely provide paired examples that reveal which dynamics should remain invariant across different morphologies. We therefore use motion abstractions not as a fixed representation for final inference, but as complementary mechanisms for constructing the cross-morphology supervision required to learn such correspondence.

Based on this insight, we propose a two-stage framework. In Stage~I, \emph{Abstract Motion Bootstrapping}, we establish a shared conditioning interface for complementary motion views, including semantic kinematics, global trajectories, dense point tracks, structural edges and 6-DoF axis. Each view captures a different aspect of motion and is applied only when its extraction is reliable. By pairing an extracted motion condition with target conditions from different categories, Stage~I synthesizes videos that differ in appearance, morphology, and background while preserving the motion attributes specified by the condition. These samples form cross-category motion-related pairs and provide explicit supervision for motion correspondence beyond a fixed structural template.

The explicit abstractions in Stage~I provide controllable supervision, but their individual coverage and information capacity remain inherently limited. Our final goal is instead to infer transferable dynamics directly from a raw source video. In Stage~II, \emph{Cross-Category Motion Internalization}, we initialize from Stage~I and replace the explicit motion condition with the source video itself. Training on cross-category pairs makes source-specific appearance and morphology unreliable predictors of the target, encouraging the model to identify dynamics that remain informative across category changes and reconstruct their realization according to the target condition. At inference, the model directly takes a source video together with either a reference image or text prompt, without requiring skeleton extraction, trajectory specification, point tracking, or per-video optimization.

We further introduce \textbf{OpenVMT-Dataset} and \textbf{OpenVMT-Bench} for training and systematic evaluation. To our knowledge, OpenVMT-Dataset is the first dedicated training dataset of open-category, cross-content video pairs with instance-level motion equivalence. It contains 10K motion-equivalent pairs spanning diverse morphologies and motion patterns, providing direct supervision for cross-category motion correspondence. OpenVMT-Bench evaluates motion transfer under progressively increasing category gaps through Same, Near, and Far splits. Our method achieves state-of-the-art performance on the proposed benchmark, demonstrating robust motion fidelity and target preservation across diverse source--target relations.

Our contributions are summarized as follows:
\begin{itemize}
    \item We formulate open-category motion transfer from the perspective of Motion Beyond Morphology, where transferable motion attributes are adaptively preserved according to the structural relationship between source and target.
    \item We propose a two-stage framework that leverages complementary motion abstractions to bootstrap cross-category supervision and subsequently internalizes this supervision into direct source-video-conditioned motion transfer.
    \item We introduce OpenVMT-Dataset, the first open-category training dataset with instance-level motion-equivalent cross-content pairs, and OpenVMT-Bench, on which our method achieves state-of-the-art performance.
\end{itemize}

\section{Related Work}

\paragraph{Structure-conditioned animation.}
Most motion-transfer methods have been developed for humans, portraits, and articulated characters. Pose-guided animation methods~\cite{ma2024followyourpose,hu2024animateanyone,xu2023magicanimate,fang20263dawareimplicitmotioncontrol,kling-motion-control} obtain precise control from keypoints, skeletons, or dense pose. Their structural assumptions are useful within a category but do not extend naturally to arbitrary objects. The same limitation applies to landmark-based face and body animation, where the motion signal is defined on a fixed morphology.

\paragraph{Reference-based motion transfer.}
Recent diffusion-based methods reduce reliance on category-specific structures by extracting motion directly from reference videos~\cite{gao2025conmo,ling2024motionclone,yesiltepe2024motionshop,ditflow,dismo}. MotionClone~\cite{ling2024motionclone}, DiTFlow~\cite{ditflow}, and MotionShop~\cite{yesiltepe2024motionshop} leverage temporal attention, motion flow, or diffusion guidance for motion transfer, while DeT~\cite{det} and DisMo~\cite{dismo} further explore motion--appearance decoupling and transferable motion representations. Despite broader applicability, such motion cues may still retain source-specific shape, layout, or appearance. Our method instead uses complementary explicit motion abstractions to construct cross-category supervision, which is subsequently internalized into direct source-video-conditioned transfer.

\paragraph{Motion representations and evaluation.}
Trajectory-based methods~\cite{yin2023dragnuwa,wang2024motionctrl, shi2024motion} control coarse motion, track-based methods~\cite{zhou2025trackgo} capture finer local dynamics, and pose-based animation~\cite{hu2024animateanyone,xu2023magicanimate} provides semantic structure for articulated subjects. Hybrid approaches~\cite{wang2023videocomposer} combine multiple cues. We instead use complementary motion abstractions according to their applicability across diverse objects.

General benchmarks such as VBench~\cite{huang2024vbench} and VBench++~\cite{huang2024vbenchpp} evaluate overall video quality, while MotionBench~\cite{yesiltepe2024motionshop} and MTBench~\cite{det} target motion transfer. Existing benchmarks offer limited control over structural gaps and rarely include curated target images. OpenVMT-Bench supports T2V and I2V with \emph{Same}, \emph{Near}, and \emph{Far} splits for motion fidelity, target preservation, and source leakage.

\section{Method}
\label{sec:method}

\begin{figure*}[t]
    \centering
    \includegraphics[width=\linewidth]{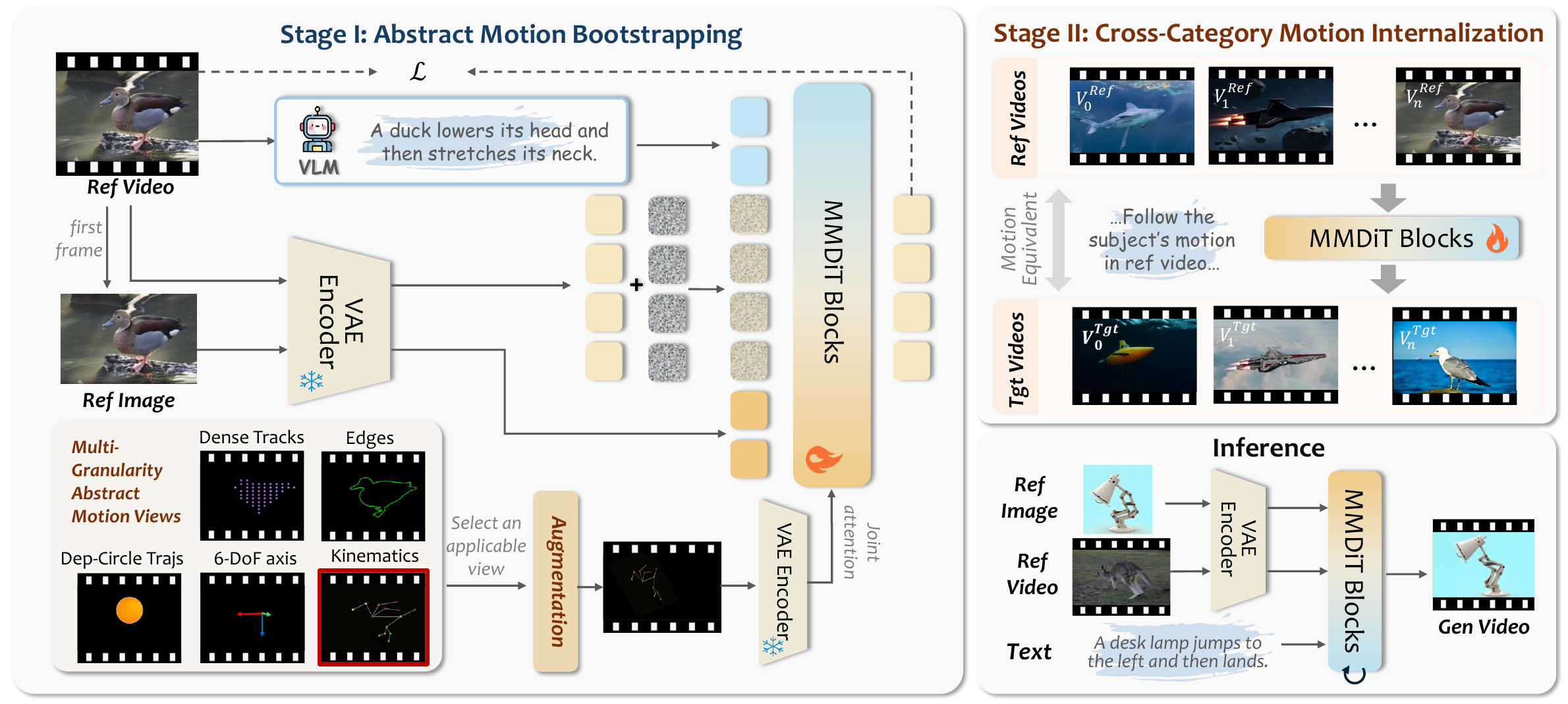}
    \caption{
    \textbf{Overview of our framework.} Left: Stage~I learns heterogeneous abstract motion conditions through a unified video-like interface and bootstraps filtered cross-category motion pairs. Upper right: Stage~II internalizes this supervision by replacing abstract motion conditions with raw reference videos. Lower right: inference directly conditions on a reference video, text, and an optional reference image, without explicit motion extraction or test-time optimization.
    }
    \label{fig:framework}
\end{figure*}

\subsection{Overview}
\label{sec:method_overview}

Given a reference video $V_r$ and a target condition $c_t$, our goal is to synthesize
\begin{equation}
    \hat{V}_t = \mathcal{G}(V_r,c_t),
\end{equation}
such that $\hat{V}_t$ follows the transferable dynamics of $V_r$, while its appearance, morphology, and scene content are determined by $c_t$.
Here, $c_t$ comprises a target text description and an optional reference image, covering both T2V and I2V settings.
Our framework follows a two-stage training scheme.
\textbf{Stage I} learns motion-conditioned generation from multi-granularity abstract motion views and uses the learned model to bootstrap cross-category video pairs with shared dynamics but different appearances and morphologies.
Since these abstract views are representation-specific and require explicit extraction, Stage I serves primarily to construct cross-category supervision rather than as the final transfer model.
\textbf{Stage II} initializes from Stage I and replaces the abstract condition with the raw reference video, thereby internalizing transferable motion into direct video-conditioned generation.
At inference, motion transfer is performed directly from the reference video under text and optional image conditioning, without explicit motion extraction or test-time optimization.

\subsection{Multi-Granularity Abstract Motion Views}
\label{sec:motion_views}

No single representation can reliably cover the diverse objects and motion patterns in open-category motion transfer.
We therefore employ multi-granularity abstract motion views, each capturing complementary motion attributes and applied only to the data for which it can be reliably extracted.

\paragraph{Semantic kinematics.}
We use ViTPose++~\cite{xu2023vitpose++} to extract animal keypoints and render skeleton sequences, providing semantically meaningful control for articulated motions such as limb movement and wing articulation.

\paragraph{Depth-aware global trajectories.}
For object-level motion, we track a stable object center derived from foreground masks and render a depth-aware circle sequence using the tracked position and mean foreground depth.
This representation captures global displacement together with approaching and receding dynamics while largely suppressing object-specific structure.

\paragraph{Dense point tracks.}
We extract dense trajectories with TAPNext++~\cite{jung2026tapnext++}, which capture local motion and deformation without predefined structural semantics, making them applicable to diverse non-rigid objects.

\paragraph{6-DoF axis.}
For rigid or approximately rigid objects, we estimate frame-wise rigid transformations from the 3D tracks produced by SpaTrackerV2~\cite{xiao2025spatialtrackerv2} and render them as coordinate-axis sequences, explicitly representing turning and orientation changes.

\paragraph{Edges.}
We further use foreground Canny edges to preserve coarse shape evolution while suppressing most appearance and background information, providing a broadly applicable structural motion cue.

These motion views provide complementary supervision across diverse motion regimes. View applicability is defined over 100+ object categories and motion patterns, with one applicable view sampled per video. After view-specific perturbations, all views are unified into a video-like form, yielding $\mathcal{D}_{\mathrm{abs}}={(M_i^k,c_i,V_i)}$. 



\subsection{Stage I: Abstract Motion Bootstrapping}
\label{sec:stage1}

\paragraph{Abstract motion grounding.}
Stage~I grounds heterogeneous motion abstractions into the generative model, enabling them to serve as controllable motion conditions.
For each training tuple $(M_i^k,c_i,V_i)\sim\mathcal{D}_{\mathrm{abs}}$, $c_i$ consists of the first frame of $V_i$ and its VLM-generated caption.
After representation-specific augmentation, the motion view $M_i^k$ and the first-frame image are encoded by the pretrained 3D VAE, while the caption is processed by the text encoder.
These conditions are incorporated with the noisy target latent through the backbone's native conditioning interface, and the model is trained to reconstruct $V_i$.
This formulation allows a single generator to learn heterogeneous motion abstractions without representation-specific control branches.

Following the flow-matching objective of the base model, we optimize
\begin{equation}
    \mathcal{L}_{\mathrm{S1}}
    =
    \mathbb{E}
    \left[
        \left\|
            u_\tau -
            u_{\theta_1}
            \left(
                z_\tau,\tau,z_m,c_i
            \right)
        \right\|_2^2
    \right],
\end{equation}
where $z_m=\mathcal{E}(M_i^k)$, and $z_\tau$ and $u_\tau$ denote the noisy latent of $V_i$ and the corresponding target velocity, respectively.

\paragraph{Cross-category motion-pair generation.}
Once grounded, the abstract views provide an intermediate motion interface for synthesizing dynamics across different visual contents.
Given a reference video $V_r$, we extract an applicable motion view
\begin{equation}
    M_r^k=\Phi_k(V_r),
\end{equation}
and pair it with a target condition $c_t$ selected from a different category.
The Stage-I model then generates
\begin{equation}
    \tilde{V}_t
    =
    \mathcal{G}_{\theta_1}
    \left(
        M_r^k,c_t
    \right).
\end{equation}
This yields candidate triplets $(V_r,c_t,\tilde{V}_t)$, where the reference and target differ in appearance and morphology while sharing the motion attributes conveyed by $M_r^k$.
In this way, Stage~I converts readily extractable motion abstractions into cross-category supervision that is difficult to obtain from real videos directly.

\paragraph{Cross-category pair filtering.}
To ensure the reliability of the bootstrapped supervision, we perform quality control both before and after synthesis.
We first reject samples with unreliable motion extraction using representation-specific validity checks.
For each generated triplet $(V_r,c_t,\tilde{V}_t)$, we further assess four complementary criteria: \emph{motion fidelity}, \emph{target fidelity}, \emph{reference leakage}, and \emph{video quality}.
These criteria respectively evaluate whether $\tilde{V}_t$ preserves the intended dynamics, follows the target condition, avoids inheriting reference-specific appearance or structure, and maintains satisfactory visual and temporal quality. 
Only candidates satisfying all criteria are retained, forming the filtered bootstrapped dataset $\mathcal{D}_{\mathrm{boot}}$ used in Stage~II.

\begin{table*}[t!]
\centering
\caption{
Quantitative comparison on OpenVMT-Bench.
We evaluate \textit{motion fidelity}, \textit{target fidelity},
\textit{leakage}, and \textit{temporal quality}.
HMF is reported across Same, Near, and Far category transfers,
while the remaining metrics are evaluated over the full benchmark.
G-Mot. and G-Leak denote Gemini-based motion preservation and leakage scores, respectively. The best and second-best results are highlighted in
\textbf{boldface} and \underline{underlining}.
}
\label{tab:quantitative}
\footnotesize
\setlength{\tabcolsep}{3.2pt}
\renewcommand{\arraystretch}{1.05}

\begin{tabular*}{0.9\textwidth}{
@{\extracolsep{\fill}}
l
ccccc
c
c
c
@{}
}
\toprule
\multirow{2}{*}{Method}
& \multicolumn{5}{c}{Motion Fidelity}
& \multicolumn{1}{c}{Target Fidelity}
& \multicolumn{1}{c}{Leakage}
& \multicolumn{1}{c}{Temporal} \\
\cmidrule(lr){2-6}
\cmidrule(lr){7-7}
\cmidrule(lr){8-8}
\cmidrule(lr){9-9}

& HMF-S $\uparrow$
& HMF-N $\uparrow$
& HMF-F $\uparrow$
& HMF-All $\uparrow$
& G-Mot. $\uparrow$
& Align. $\uparrow$
& G-Leak $\downarrow$
& Smooth. $\uparrow$ \\

\midrule

\multicolumn{9}{l}{
\textit{OpenVMT-I2V}
\quad
\small Align.: DINO-I $\uparrow$
} \\[-1pt]

DisMo
& 0.7367
& \underline{0.7437}
& \underline{0.7441}
& \underline{0.7412}
& \underline{2.811}
& 0.8587
& 1.182
& 0.8018 \\

Wan-Move
& \underline{0.7370}
& 0.7338
& 0.7335
& 0.7349
& 2.019
& \underline{0.8849}
& 1.128
& \underline{0.9498} \\

Tora
& 0.7161
& 0.7259
& 0.7402
& 0.7262
& 2.377
& 0.8846
& \underline{1.023}
& 0.9310 \\

\addlinespace[1pt]
\textbf{Ours}
& \textbf{0.7435}
& \textbf{0.7617}
& \textbf{0.7446}
& \textbf{0.7505}
& \textbf{3.575}
& \textbf{0.9058}
& \textbf{1.000}
& \textbf{0.9651} \\

\midrule

\multicolumn{9}{l}{
\textit{OpenVMT-T2V}
\quad
\small Align.: CLIP-T $\uparrow$
} \\[-1pt]

DeT
& 0.7441
& 0.7486
& 0.7358
& 0.7428
& 2.418
& 0.2392
& 1.152
& 0.9031 \\

FlowMotion
& 0.7384
& 0.7440
& 0.7337
& 0.7387
& \underline{2.970}
& \underline{0.2522}
& 1.170
& 0.8859 \\

DisMo-T2V
& \underline{0.7680}
& \underline{0.7666}
& \textbf{0.7742}
& \underline{0.7696}
& 2.873
& 0.2415
& \underline{1.109}
& \textbf{0.9265} \\

\addlinespace[1pt]
\textbf{Ours}
& \textbf{0.7722}
& \textbf{0.7712}
& \underline{0.7688}
& \textbf{0.7707}
& \textbf{3.976}
& \textbf{0.2677}
& \textbf{1.067}
& \underline{0.9122} \\

\bottomrule
\end{tabular*}
\end{table*}

\subsection{Stage II: Cross-Category Motion Internalization}
\label{sec:stage2}

Stage~I provides effective cross-category supervision through abstract motion views, but these representations are inherently representation-specific: they require dedicated extractors and retain only selected aspects of the underlying motion.
Our goal in Stage~II is therefore to internalize this supervision into the generator, such that transferable dynamics can be inferred directly from the raw reference video without committing to a predefined motion abstraction.

We initialize the model from Stage~I, $\theta_2 \leftarrow \theta_1$, and replace the abstract motion condition $M_r^k$ with the reference video $V_r$.
Using the same video-like conditioning interface, the reference video is encoded as
\begin{equation}
    z_r=\mathcal{E}(V_r),
\end{equation}
and conditions the generation of the bootstrapped target $\tilde{V}_t$ together with $c_t$.
For each $(V_r,c_t,\tilde{V}_t)\sim\mathcal{D}_{\mathrm{boot}}$, we optimize
\begin{equation}
    \mathcal{L}_{\mathrm{S2}}
    =
    \mathbb{E}
    \left[
        \left\|
            \tilde{u}_{\tau}
            -
            u_{\theta_2}
            \left(
                \tilde{z}_{\tau},
                \tau,
                z_r,
                c_t
            \right)
        \right\|_2^2
    \right],
\end{equation}
where $\tilde{z}_{\tau}$ and $\tilde{u}_{\tau}$ denote the noisy latent of $\tilde{V}_t$ and its corresponding target velocity.

The cross-category construction of $\mathcal{D}_{\mathrm{boot}}$ is essential to this internalization process.
Although $V_r$ and $\tilde{V}_t$ differ substantially in appearance and morphology, they preserve the motion attributes transferred through Stage~I.
Consequently, reference-specific appearance and structure no longer consistently predict the target video, whereas the shared temporal dynamics remain informative across category changes.
Training on such pairs therefore discourages reliance on morphology-specific shortcuts and encourages the model to extract transferable motion directly from RGB observations and realize it according to the target condition.

\subsection{Inference}
\label{sec:inference}

At inference, motion transfer is performed directly from the reference video,
\begin{equation}
    \hat{V}_t = \mathcal{G}_{\theta_2}(V_r,c_t),
\end{equation}
where $c_t$ denotes the target text condition with an optional reference image, covering both T2V and I2V settings.
The reference video provides the motion signal, whereas $c_t$ determines the target appearance, morphology, and scene content.
Starting from a Gaussian noise latent, the model progressively generates the target video through iterative denoising conditioned on both $V_r$ and $c_t$.

The resulting inference pipeline eliminates the need for explicit motion extraction and per-video optimization, enabling unified motion transfer across diverse target categories within a single model.

\begin{figure*}[t!]
    \centering
    \includegraphics[width=1\linewidth]{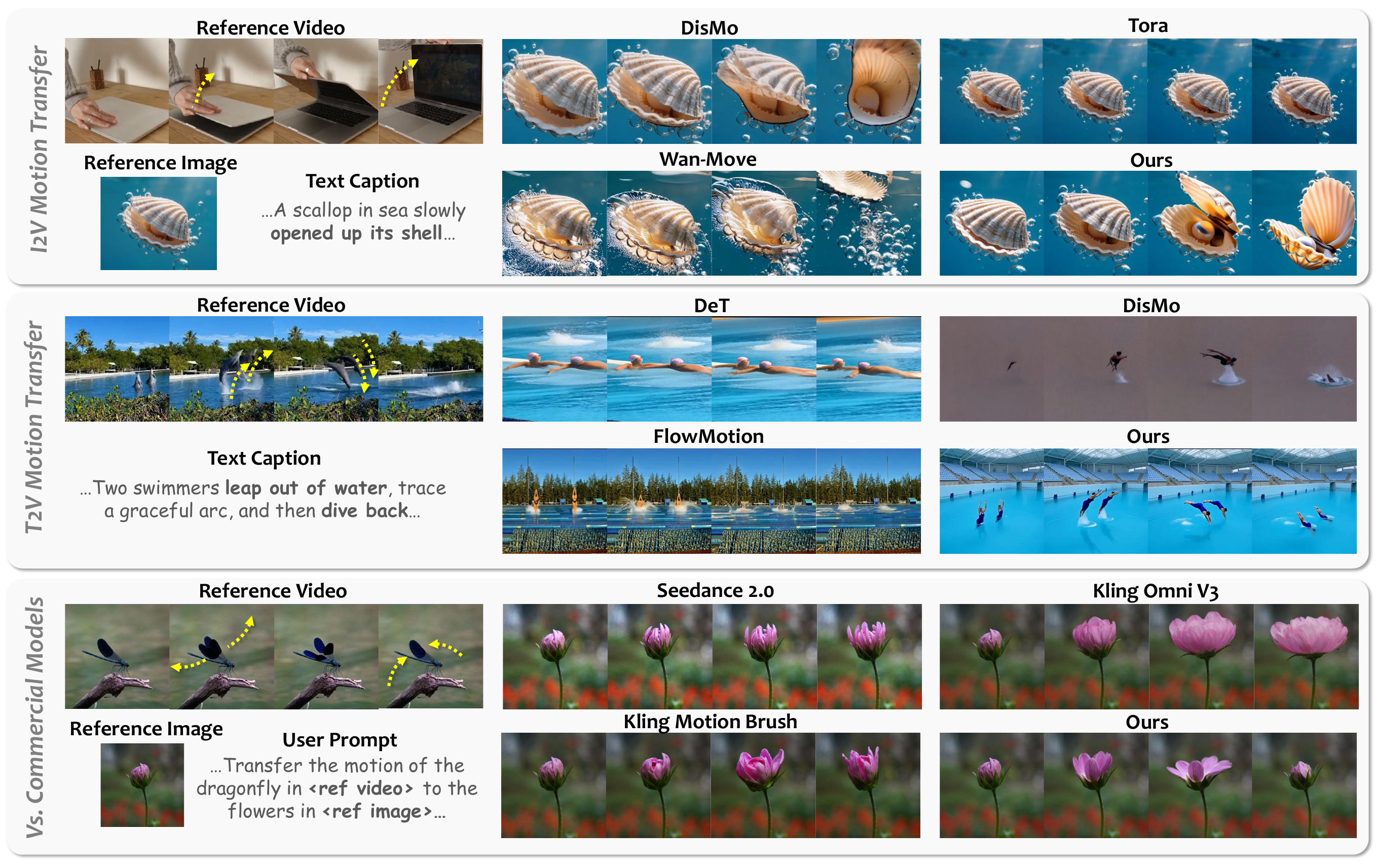}
    \caption{
Top: I2V motion transfer against open-source baselines.
    Middle: T2V motion transfer under cross-category settings.
    Bottom: comparison with commercial video generation models.
    Our method faithfully transfers diverse motion patterns, including articulated and non-rigid dynamics, while better preserving the target content across large morphology gaps.}
    \label{fig:qualitative_comparison}
\end{figure*}

\section{Experiments}
\label{sec:experiments}

\subsection{Experimental Setup}
\label{sec:experimental_setup}

\paragraph{Dataset and benchmark.}
We train our model on a large-scale collection of approximately 1.6M object-centric motion videos curated from the Internet, covering diverse subjects and motion patterns. In addition, we curate OpenVMT-Dataset, a release set of 10K high-quality cross-content motion-equivalent video pairs derived from our bootstrapped data, comprising 4K Same, 4K Near, and 2K Far pairs. Videos within each pair exhibit equivalent motion dynamics while differing in subject identity, morphology, and scene background. Such paired data provides explicit supervision for learning motion beyond appearance and structural correspondence, complementing the diverse but unpaired Internet videos.

For evaluation, we introduce OpenVMT-Bench, consisting of an image-conditioned track, OpenVMT-I2V, and a text-conditioned track, OpenVMT-T2V. Each track is divided into Same, Near, and Far splits with progressively larger semantic and morphological gaps between the source motion and target content: Same pairs share closely matched categories and structures, Near pairs exhibit moderate cross-category variation, and Far pairs involve substantially different subjects with weak or absent part-level correspondence. OpenVMT-I2V and OpenVMT-T2V contain 123 and 166 test cases, respectively.

\paragraph{Baselines.}
We select representative open-source motion-transfer methods spanning different conditioning paradigms. For OpenVMT-I2V, we compare against DisMo~\cite{dismo}, which uses an implicit motion representation, and the trajectory-conditioned methods Wan-Move~\cite{wanmove} and Tora~\cite{tora}. For trajectory-based baselines, we extract foreground point trajectories using CoTracker2~\cite{Cotracker} as motion conditions.
For OpenVMT-T2V, we compare against DeT~\cite{det}, FlowMotion~\cite{flowmotion}, and a text-conditioned adaptation of DisMo (DisMo-T2V)~\cite{dismo}, obtained by disabling its image condition while retaining text and motion conditioning. We use official implementations and released checkpoints whenever available, and align output duration, resolution, and frame rate across methods. 
We additionally provide qualitative comparisons with commercial systems, including Seedance-2.0~\cite{seedance2}, Kling Motion Brush~\cite{klingai}, and Kling Omni3~\cite{kling-omni3}.

\paragraph{Metrics.}
We evaluate motion transfer along four dimensions: motion fidelity, target fidelity, source leakage, and temporal quality. Motion fidelity is measured by Hybrid Motion Fidelity (HMF)~\cite{det}, which captures consistency in global trajectories and local motion dynamics, together with a Gemini-based motion preservation score (G-Mot.) that assesses perceptual correspondence between the source and generated motions. Target fidelity is evaluated by DINO-I~\cite{oquab2024dinov2} for OpenVMT-I2V, measuring visual similarity to the reference image, and by CLIP-T~\cite{wu2024motionbooth} for OpenVMT-T2V, measuring alignment with the target text prompt. Source leakage is assessed using a Gemini-based leakage score (G-Leak), which measures the presence of source-specific appearance, structure, or background content in the generated video, with lower values indicating less leakage. Temporal quality is evaluated using the motion smoothness metric from VBench~\cite{huang2024vbench}. Since automatic metrics may not fully capture perceptual differences in motion transfer quality, we additionally conduct a human Good/Same/Bad (GSB) preference study.

\paragraph{Implementation details.}
Our model is initialized from a MMDiT-based image-to-video diffusion backbone with a causal 3D VAE.
Following the backbone training configuration, we use 121-frame video clips resized to a target area of $480\times854$ pixels while preserving the original aspect ratio.
We optimize the model using Adam with a global batch size of 128.
Stage~I is trained for 15K iterations on the large-scale unpaired motion video corpus.
Stage~II is initialized from the Stage-I model and further trained for 4K iterations on the constructed motion-equivalent pairs.

\subsection{Comparison with State-of-the-Art Methods}
\label{sec:comparison}

\paragraph{Image-to-video motion transfer.}
Table~\ref{tab:quantitative} reports results on OpenVMT-I2V. Our method achieves the best performance across all reported metrics. It consistently obtains the highest HMF from Same to Far transfers, together with a substantially higher G-Mot., demonstrating robust motion fidelity across category gaps. It also achieves the best target fidelity and lowest G-Leak, indicating stronger target preservation with less source-content contamination. The highest smoothness score further confirms that these gains are achieved without sacrificing temporal quality.

\paragraph{Text-to-video motion transfer.}
Table~\ref{tab:quantitative} reports results on OpenVMT-T2V. Our method achieves the best overall HMF, G-Mot., CLIP-T, and G-Leak, demonstrating a strong balance between motion preservation and text-conditioned target generation. While DisMo-T2V slightly outperforms ours in HMF-F and temporal smoothness, its stronger adherence to source motion is accompanied by lower target alignment and higher source leakage, indicating a tendency to preserve source-specific motion structures at the expense of faithful target realization, as shown in Fig.~\ref{fig:qualitative_comparison}. In contrast, our method substantially improves perceptual motion fidelity and text alignment while maintaining competitive motion consistency, enabling more reliable motion transfer across category gaps.

\paragraph{Perceptual evaluation.}
We further conduct a human Good/Same/Bad (GSB) preference study with 12 evaluators experienced in video generation. Since GSB evaluation is pairwise, we report results against the strongest representative baseline for each benchmark track, with DisMo used for OpenVMT-I2V and FlowMotion for OpenVMT-T2V. We report both preference rate, $\mathrm{Pref.}=(G+0.5S)/(G+S+B)$, and decisive win rate, $\mathrm{Win}=G/(G+B)$. As shown in Table~\ref{tab:human_eval}, our method is consistently preferred in motion fidelity, target preservation, and overall quality, achieving overall preference rates of \textbf{93.0\%} on I2V and \textbf{97.3\%} on T2V.

\begin{table}[t]
\centering
\caption{
Human G/S/B preference evaluation against DisMo for I2V and FlowMotion for T2V. G, S, and B denote preference for ours, tie, and preference for the baseline, respectively.
}
\label{tab:human_eval}
\footnotesize
\setlength{\tabcolsep}{3.4pt}
\renewcommand{\arraystretch}{1.00}
\begin{tabular}{llccccc}
\toprule
Task & Criterion
& G $\uparrow$
& S
& B $\downarrow$
& Pref. $\uparrow$
& Win $\uparrow$ \\
\midrule

\multirow{3}{*}{I2V}
& Motion
& 82.8 & 13.1 & 4.1
& 89.3 & 95.3 \\
& Target
& 67.2 & 24.6 & 8.2
& 79.5 & 89.1 \\
& Overall
& \textbf{88.5} & 9.0 & 2.5
& \textbf{93.0} & \textbf{97.3} \\

\midrule

\multirow{3}{*}{T2V}
& Motion
& 82.9 & 15.9 & 1.2
& 90.9 & 98.6 \\
& Target
& 77.4 & 20.7 & 1.8
& 87.8 & 97.7 \\
& Overall
& \textbf{95.7} & 3.0 & 1.2
& \textbf{97.3} & \textbf{98.7} \\

\bottomrule
\end{tabular}

\vspace{1pt}
\end{table}


\subsection{Qualitative Comparison}
\label{sec:qualitative_comparison}
Figure~\ref{fig:qualitative_comparison} presents qualitative comparisons on I2V (top), T2V (middle), and commercial video generators (bottom). For I2V, trajectory-based methods mainly capture coarse displacement, while DisMo may retain source-specific structures under large morphology gaps. Our method better preserves fine-grained motion while maintaining the target appearance and structure. For T2V, it more faithfully transfers both global and local dynamics while preserving the target semantics. Compared with commercial models, including Seedance-2.0~\cite{seedance2}, Kling Motion Brush~\cite{klingai}, and Kling Omni3~\cite{kling-omni3}, our method shows stronger reference-motion adherence across large morphology gaps, faithfully reproducing the specific dynamics of the source video rather than merely plausible target motion.

\begin{table}[t]
\centering
\caption{
Ablation of cross-category bootstrapping and motion abstractions on OpenVMT-T2V.
}
\label{tab:ablation}
\footnotesize
\setlength{\tabcolsep}{5.0pt}
\renewcommand{\arraystretch}{1.08}

\begin{tabularx}{\columnwidth}{
@{}
>{\raggedright\arraybackslash}X
ccc
@{}
}
\toprule
Variant
& HMF-All $\uparrow$
& HMF-F $\uparrow$
& G-Mot. $\uparrow$ \\
\midrule

\multicolumn{4}{@{}l}{\textit{Stage I: Multi-granularity motion views}} \\

Trajectory only
& 0.7524
& 0.7582
& 3.354 \\

Track only
& 0.7593
& 0.7575
& 3.673 \\

Edge only
& 0.7656
& 0.7638
& 3.646 \\

\midrule
\multicolumn{4}{@{}l}{\textit{Stage II: Cross-category bootstrapping}} \\

In-category bootstrap
& 0.7588
& 0.7538
& 3.436 \\

\midrule
\textbf{Full model}
& \textbf{0.7707}
& \textbf{0.7688}
& \textbf{3.976} \\

\bottomrule
\end{tabularx}
\end{table}

\begin{figure}
    \centering
    \includegraphics[width=1\linewidth]{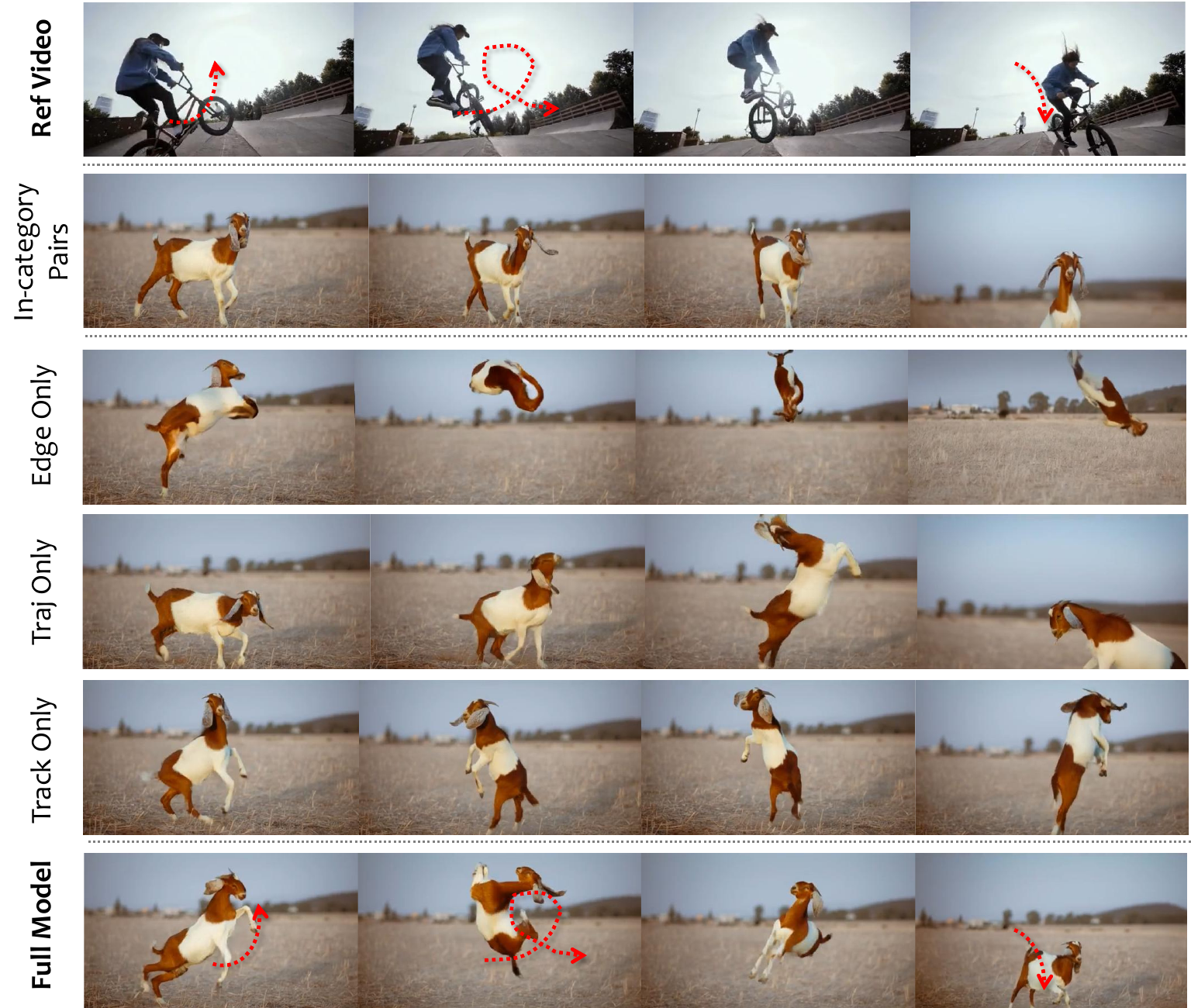}
    \caption{Qualitative ablation of cross-category bootstrapping and motion views. The full model better balances global trajectory, local motion details, pose, and target preservation.}
    \label{fig:ablation}
\end{figure}

\subsection{Ablation studies}
Table~\ref{tab:ablation} reports quantitative ablations on OpenVMT-T2V, while Fig.~\ref{fig:ablation} shows qualitative results on OpenVMT-I2V. Replacing cross-category pairs in Stage~II with in-category pairs reduces HMF-All from 0.7707 to 0.7588 and G-Mot. from 3.976 to 3.436, confirming the benefit of morphology-diverse supervision.

We further analyze individual motion views with Stage~I to characterize their representational behavior. Trajectories favor global displacement, tracks capture local dynamics, and edges provide stronger structural cues, highlighting their complementary roles across motion regimes.


\section{Conclusion}
\label{sec:conclusion}
We presented Motion Beyond Morphology, a two-stage framework for open-category video motion transfer.
The central idea is to use heterogeneous motion abstractions to construct cross-category supervision rather than imposing a fixed representation at inference.
Stage~I grounds semantic kinematics, depth-aware global trajectories, dense point tracks, 6-DoF axis, and structural edges through a shared conditioning interface, and uses them to bootstrap motion-equivalent videos across diverse morphologies.
Stage~II internalizes this supervision into direct source-video-conditioned generation, supporting both image- and text-conditioned motion transfer without explicit motion extraction at inference.  Experiments on OpenVMT-Bench examine motion fidelity, target preservation, and source-content leakage across increasingly difficult category gaps.


\bibliography{aaai2027}


\end{document}